\documentclass{article}

\usepackage[final]{main}
\usepackage{multirow}

\usepackage{graphicx}
\usepackage[utf8]{inputenc} 
\usepackage[T1]{fontenc}    
\usepackage{hyperref}       
\usepackage{url}            
\usepackage{booktabs}       
\usepackage{amsfonts}       
\usepackage{nicefrac}       
\usepackage{microtype}      
\usepackage{xcolor}        

\usepackage{cuted}
\usepackage{capt-of}
\usepackage{pgfplots}
\usepackage{colortbl}
\usepackage{graphicx}
\usepackage{subcaption}
\usepackage{enumitem}
\usepackage{multicol}
\usepackage{multirow}
\usepackage{float}
\usepackage{wrapfig}

\usepackage{amsmath}
\usepackage{amssymb}

\definecolor{blue}{HTML}{BDD8FF}

\title{A Pixel Is Worth More Than One 3D Gaussians in Single-View 3D Reconstruction}

\author{%
  Jianghao Shen$^1$ \quad Nan Xue$^2$ \quad Tianfu Wu$^1$ \\
  $^1$Department of ECE, NC State University \quad $^2$Ant Group\\
  \texttt{\{jshen27,tianfu\_wu\}@ncsu.edu} \\
  \texttt{xuenan@ieee.org}
  }

\begin{document}

\maketitle

\begin{abstract}
  Learning 3D scene representation from a single-view image is a long-standing  fundamental problem in computer vision, with the inherent ambiguity in predicting contents unseen from the input view. Built on the recently proposed 3D Gaussian Splatting (3DGS), the Splatter Image method has made promising progress on fast single-image novel view synthesis via learning a single 3D Gaussian for each pixel based on the U-Net feature map of an input image. However, it has limited expressive power to represent occluded components that are not observable in the input view. To address this problem, this paper presents a Hierarchical Splatter Image method in which \textit{a pixel is worth more than one 3D Gaussians}. Specifically, 
  each pixel is represented by a \textit{parent} 3D Gaussian and a small number of \textit{child} 3D Gaussians. Parent 3D Gaussians are learned as done in the vanilla Splatter Image. Child 3D Gaussians are learned via a lightweight Multi-Layer Perceptron (MLP) which takes as input the projected image features of a parent 3D Gaussian and the embedding of a target camera view. Both parent and child 3D Gaussians are learned end-to-end in a stage-wise way. The joint condition of input image features from eyes of the parent Gaussians and the target camera position facilitates learning to allocate child Gaussians to ``see the unseen'', recovering the occluded details that are often missed by parent Gaussians.
  In experiments, the proposed method is tested on the ShapeNet-SRN and CO3D datasets with state-of-the-art performance obtained, especially showing promising capabilities of reconstructing occluded contents in the input view. 
\end{abstract}

\setlength{\abovecaptionskip}{0pt}
\setlength{\belowcaptionskip}{-6pt}
\setlength{\belowdisplayskip}{1pt} \setlength{\belowdisplayshortskip}{1pt}
\setlength{\abovedisplayskip}{1pt} \setlength{\abovedisplayshortskip}{1pt}

\section{Introduction}
\vspace{-2mm}
\begin{figure} [h]
    \centering
    \includegraphics[width=0.9\linewidth]{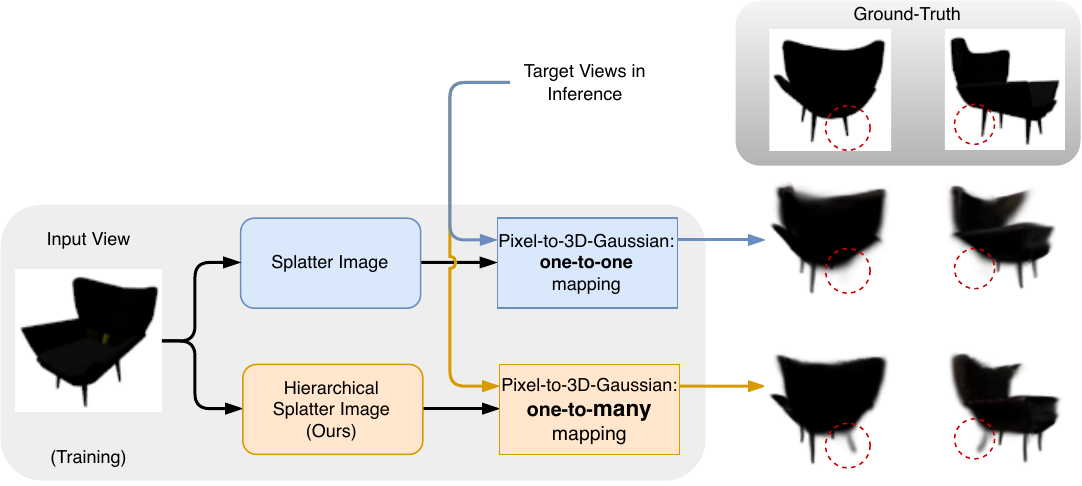}
    \caption{\small 
    Illustration of the proposed Hierarchical Splatter Image in comparison with the vanilla Splatter Image~\cite{szymanowicz2023splatter}. The former is built on the latter. The main difference lies in their different answers to the question of how many 3D Gaussians a pixel is worth in learning single-view image 3D reconstruction. Our proposed method generalizes the one-to-one pixel-to-3D-Gaussian mapping utilized in the vanilla Splatter Image to an one-to-many mapping with a two-layer hierarchical parent-child 3D Gaussian representation. We show the proposed method can sensibly recover occluded parts (e.g. chair legs). The images are from the ShapeNet-SRN dataset~\cite{chang2015shapenet}.  See text for details.
    }
    \label{fig:teaser} \vspace{-3mm}
\end{figure}

Single-view image 3D reconstruction is a long-standing problem in computer vision, with applications ranging from robotic navigation to augmented reality. One fundamental challenge of the problem lies in the intrinsic ambiguity in extracting 3D scene information (e.g., geometry and appearance) from a single 2D image. In the early prior art, this problem has been posed as a monocular depth prediction/regression  problem ~\cite{zhou2017unsupervised,zhan2018unsupervised,yuan2203new,watson2019self,shu2020feature,luo2019every,guizilini20203d,gonzalezbello2020forget,godard2019digging} which learns the pixel-to-depth mapping in a view-specific way and can not naturally and elegantly support novel view synthesis.

To tackle the limitation of the pixel-to-depth regression, pioneered by the Neural Radiance Field (NeRF) method ~\cite{mildenhall2021nerf}, different NeRF-based methods have recently been proposed to learn a more holistic 3D scene representation that naturally supports novel view synthesis. NeRF is a volumetric representation representing a scene by learning a scene-specific coordinate  Multi-Layer Perceptron (MLP) from multi-view images (which are often entailed to be densely sampled with ground-truth camera poses known). The novel view synthesis/rendering process is done by ray marching with the learned coordinate MLP for each pixel in a target view. 
 
Sparse/single view based NeRFs ~\cite{yu2021pixelnerf,jang2021codenerf,lin2023vision,gu2023nerfdiff} have been proposed to relax the requirements in the vanilla NeRF for more general applications, where coordinate MLPs are further conditioned on latent codes embedded with scene level prior information. 

Although NeRF-based methods enable high-fidelity novel view synthesis, their implicit coordinate-MLP design and explicit dense-point ray marching based rendering lead to high computational demands in both training and inference. 

More recently, 3D Gaussian Splatting (3DGS) ~\cite{kerbl20233d} has been proposed  as an alternative 3D scene representation to NeRF, which explicitly represents a scene by learning a mixture of, often a sufficiently large number of, 3D Gaussians through a highly adaptive density control scheme, which in turn enables real-time rendering through the 3D-to-2D splatting operation. The vanilla 3DGS also needs relatively dense multi-view posed images as inputs and the learned 3DGS is scene specific. 

To extend the vanilla 3DGS for single-view 3D reconstruction, the Splatter Image~\cite{szymanowicz2023splatter} method proposes to learning an one-to-one pixel-to-3D-Gaussian mapping using an U-Net type image encoder, and utilizes the learned 3DGS of all pixels for novel view synthesis in an ultra-fast way following the vanilla 3DGS rendering procedure, which has shown promising performance and is the state-of-the-art method. 

However, in testing, Splatter Image still struggles to reliably and accurately reconstruct target views that are significantly different from the input view, in particular, those with occluded structures that are not visible in the input image (see chair legs in Fig.~\ref{fig:teaser}). 
Based on our analyses, this problem seems to be caused by two issues: \textit{First}, Splatter Image only predicts a single 3D Gaussian for each pixel, which limits its representation power to reconstruct views that have complex structures; \textit{Second}, the prediction of 3D Gaussians in Splatter Image is based on input image features alone, which does not have sufficient conditional information to guide the Gaussians in representing  structures that are not visible in the input view, resulting in either blurry artifacts, or missing occluded components completely when rendering a novel target view.

\begin{figure} [t]
    \centering
    \includegraphics[width=0.7\linewidth]{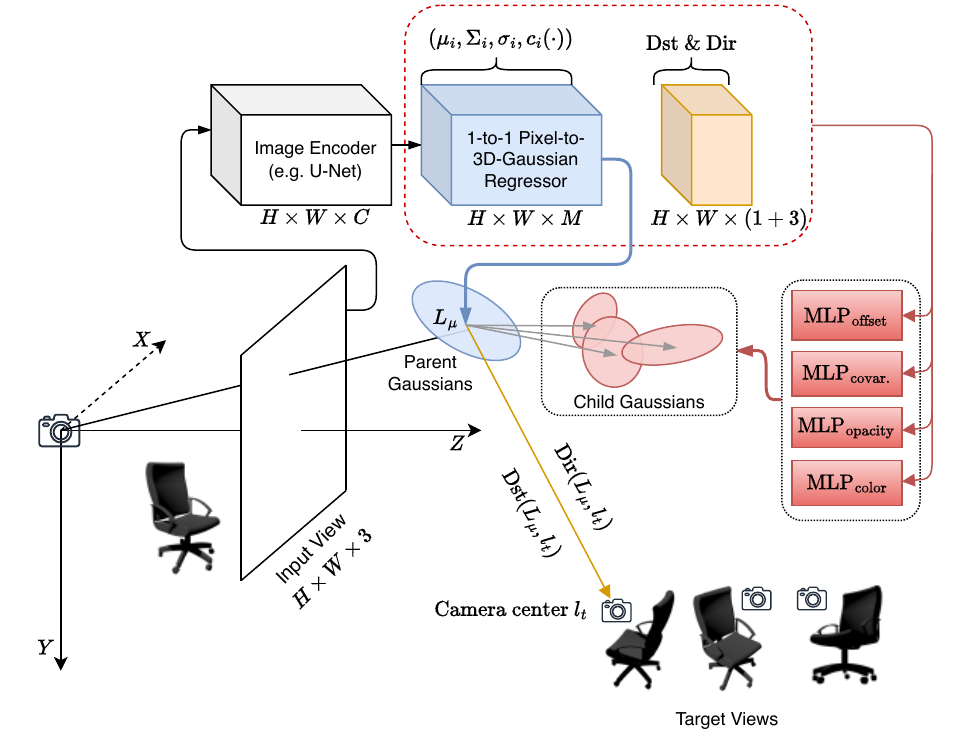}
    \caption{\small Illustration of the proposed Hierarchical Splatter Image which is built on, and aims to address two issues of, the vanilla Splatter Image~\cite{szymanowicz2023splatter}. During training, we estimate the parameters of the image encoder, parent 3D Gaussian regressor and MLPs using the rendering loss between the 3DGS rendered images using the learned parent-child 3D Gaussian and the ground-truth images. During inference, we have an input view of an object instance that is unseen during training (e.g., the input in Fig.~\ref{fig:teaser}), we first compute the parent Gaussians. Then, based on a target view of interest, we compute the child Gaussians via MLPs. With all the 3D Gaussians computed, we can synthesize the image via 3DGS rendering. In comparisons, the vanilla Splatter Image computes the parent Gaussian only and renders images for any target views. Our method entails executing MLPs on top of the parent Gaussians for each target view. We show this overhead is negligible in terms of FLOPs.   See text for details. }
    \label{fig:method} \vspace{-2mm}
\end{figure}

In this paper, we aim to directly address the above two issues of the vanilla Splatter Image method. As illustrated in Fig.~\ref{fig:method}, we propose to learn more than one 3D Gaussians per pixel to induce overcompleteness to facilitate better representational expressivity. More specifically, we present a method of learning a two-layer hierarchy of parent-child 3D Gaussians at each pixel in the single-view input image. The parent 3D Gaussians are the same as those in the vanilla Splatter Image. The key idea of this paper is how the child 3D Gaussians are learned to be more structure-aware, such that they will be capable of recovering parts which are visible from a target view in inference, but occluded in the input view. The child 3D Gaussians are learned conditioning on the parent 3D Gaussian and target views. More specifically, child 3D Gaussians are learned via a lightweight  MLP which takes as input the corresponding image features of the  parent 3D Gaussian and the embedding of a target camera view. 
When trained under photometric reconstruction loss, the joint condition of input image features and target view embeddings guide our model to allocate child 3D Gaussians into positions that require view-specific refinement, thus the missing structural details of occluded areas can be retrieved (see the two examples of chair legs in Fig.~\ref{fig:teaser}).

\textbf{Our Contributions.} This paper makes three main contributions to the field of single-view 3D reconstruction: 
(i) It presents Hierarchical Spatter Image which generalizes the one-to-one pixel-to-3D-Gaussian mapping in the vanilla Splatter Image to the one-to-many pixel-to-3D-Gaussian mapping. The one-to-many mapping is organized into a hierarchy of parent-child 3D Gaussians. (ii) It proposes a simple yet effective method for jointly learning parent and child 3D Gaussians to induce better structural awareness. The structural awareness is achieved by explicitly encoding the spatial relationships between parent 3D Gaussians and target camera poses via lightweight MLPs. (iii) It achieves state of the art performance in four single image 3D reconstruction tasks (Chairs and Cars in the ShapeNet-SRN benchmark, and Hydrants and Teddybears in the CO3D benchmark). In particular, it reconstructs occluded contents with higher accuracy than Splatter Image.

\vspace{-2mm}
\section{Approach}
\vspace{-2mm}

In this section, we first define the problem of single-view 3D reconstruction studied in this paper, and give an overview of the Splatter Image~\cite{szymanowicz2023splatter} method. We then present details of the proposed Hierarchical Splatter Image.  

\vspace{-2mm}
\subsection{Problem Definition of Single-View 3D Reconstruction Using 3DGS}
\vspace{-2mm}

We adopt the category-specific pixel-to-3D-Gaussian mapping pipeline. Denote by $D^{\text{train}}=\{ (I_{i,j}, \mathcal{P}_{i,j}); i=1,\cdots n \}$ the training set of images consisting of multi-view object instances of the same class (e.g., chair or car in ShapeNet-SRN~\cite{chang2015shapenet}), where $I_{i,j}$ represents the $j$-th view RGB image of the object instance $i$ with a shape of $H\times W\times 3$, and $\mathcal{P}_{i,j}$ its camera pose (both instrinsic matrix and extrinsic matrix). There are a large variation of appearance and structures among different object instances, which poses a significant challenge for learning reliable single-view 3D reconstruction. In this paper, we assume there are at least two views of a same object instance  in the training dataset, one will be used as the input view and the other as target view. If there are multiple views available, we will randomly sample a small subset of views (e.g.,  4 following~\cite{szymanowicz2023splatter}) and use one of them as the input view and the remaining as targete views (see Fig.~\ref{fig:method}). Similarly, $D^{\text{test}}$ can be defined and $D^{\text{train}}\cap D^{\text{test}}=\emptyset$ which means there are no overlapping object instances in the two sets.

In the 3DGS~\cite{kerbl20233d} a 3D Gaussian is represented by a 4-tuple,
\begin{equation}
    \theta_i = (\mu_i, \Sigma_i, \sigma_i, c_i(\cdot)), \label{eq:3d-gaussian}
\end{equation}
where $\mu_i\in \mathbb{R}^3$ is the Gaussian mean, $\Sigma_i$ its $3\times 3$ covariance matrix defining the shape and size, $\sigma_i\in \mathbb{R}_+$ its opacity, and $c_i(v)\in \mathbb{R}^{SH}$ a view-dependent color using spherical harmonics (SH), e.g., $SH=12$. To ensure positive semi-definiteness of the covariance matrix during optimization, the 3D covariance matrix $\Sigma$ is decomposed into two learnable components: the quaternion $R\in \mathbb{R}^4$ representing rotation, and the 3D vector $S\in \mathbb{R}^3$ representing scaling along the $X,Y$ and $Z$ axes. The  3D covariance matrix $\Sigma$ is rewritten by,
\begin{equation}
    \Sigma = RSS^{\top}R^{\top}.
\end{equation}

Our goal of single-view 3D reconstruction is to infer a mixture of 3D Gaussians from a single-view input image $(I, \mathcal{P})$,
\begin{equation}
    \mathcal{G}(I,\mathcal{P})=(\theta_1, \theta_2, \cdots, \theta_G), \label{eq:3dgs_srn}
\end{equation}
with which we can synthesize the image  of the object in $I$ for a given target view  $\mathcal{P}'$ (see examples in Fig.~\ref{fig:teaser}), $\hat{I}(\mathcal{P}')=\mathcal{R}(\mathcal{G}(I, \mathcal{P}), \mathcal{P}')$ using the fast differentiable renderer $\mathcal{R}(\cdot)$ provided by 3DGS~\cite{kerbl20233d}. 

The challenge of category-level single-view 3D reconstruction using 3DGS lies in how to parameterize and learn the mixture of 3D Gaussian, $\mathcal{G}(I, \mathcal{P})$, to induce high-fidelity novel view synthesis, especially for target views $\mathcal{P}'$'s that are significantly different from the input view $\mathcal{P}$.

\vspace{-2mm}
\subsection{Background on Splatter Image}
\vspace{-2mm}

Splatter Image~\cite{szymanowicz2023splatter} presents an effective  framework which directly infers a 3D Gaussian (Eqn.~\ref{eq:3d-gaussian}) for each pixel in the input view $I$. To that end, it parameterizes $\mathcal{G}(I,\mathcal{P})$ (Eqn.~\ref{eq:3dgs_srn}) using two components: 
\begin{itemize} [leftmargin=*, noitemsep]
    \item An image encoder, denoted by $f(\cdot;\phi)\in \mathbb{R}^{H\times W\times C}$, which computes the $C$-dim feature map of the same shape as the input image by utilizing a U-Net type feature backbone. $\phi$ collects the parameters of the encoder. 
    \item An pixel-wise 3D Gaussian regression head, denoted by $g(\cdot;\psi)\in \mathbb{R}^{H\times W\times M}$, which computes $\theta$'s (Eqn.~\ref{eq:3d-gaussian}) at each pixel in the encoded feature map $f(I;\phi)$ by a simple $1\times 1$ convolutional layer. Some of the $M$ channels in the output will go through certain nonlinearity functions (e.g., the Sigmoid function for the opacity channel). 
\end{itemize}
So, Splatter Image has a straightforward one-to-one pixel-to-3D-Gaussian parameterization scheme for Eqn.~\ref{eq:3dgs_srn}, 
\begin{equation}
    \mathcal{G}(I, \mathcal{P}) = g(f(I;\phi);\psi) = (\theta_1, \cdots, \theta_{H\times W}). \label{eq:splatter-image}
\end{equation}

Instead of directly regressing the mean $\mu$ of 3D Gaussian (which may lead to unstable training due to the well-known challenge of 2D-to-3D lifting), and to account for the spatial flexibility of the correspondence between a 2D pixel and its 3D Gaussian, Splatter Image utilizes a reparameterization trick. For each pixel $(x, y, 1)$ (using the homogeneous coordinate system), it predicts the depth $d$ and the 3D offset vector $(\delta X, \delta Y, \delta Z)$, and the computes the mean by,
\begin{equation}
    \mu = \begin{bmatrix}
        x\cdot d + \delta X \\
        y\cdot d + \delta Y \\
         d + \delta Z 
    \end{bmatrix}.
\end{equation}

In sum, the output of Splatter Image is a $H\times W\times M$  dimensional tensor with $M=24$, consisting of $(\hat{d}, \delta X, \delta Y, \delta Z; R_{4\times 1}, S_{3\times 1}; \hat{\sigma}; c_{12\times 1})$. 
The actual predicted depth is then computed by $d=(z_{\text{far}}-z_{\text{near}})\cdot \text{Sigmoid}(\hat{d}) + z_{\text{near}}$ where $z_{\text{near}}$ and $z_{\text{far}}$ are predefined scene range of interest. The opacity is computed by $\sigma=\text{Sigmoid}(\hat{\sigma})$. 
Please refer to~\cite{szymanowicz2023splatter} for more details of Splatter Image.

\textbf{Limitations of Splatter Image.} From above, we see that Splatter Image predicts both depth value and spatial offsets for each 3D Gaussian. One underlying benefit of doing so is that some of the Gaussians represent visible parts/regions of the scene, and others can move freely to represent unseen parts/regions of the scene. However, since the input view only contains contextual cues of visible parts/regions, there is not enough conditional guidance for Splatter Image to infer Gaussians that can cover target views that are significantly different from the input view. In addition, Splatter Image uses a fixed number of 3D Gaussians to represent the entire scene, which may suffer from limited view-dependent interpolation capabilities, and thus makes the rendering less structurally reliable. Thus, Splatter Image tends to render images with blurry artifacts or missing geometric details for views that are substantially different from the input (Fig.~\ref{fig:teaser}). 

\subsection{Our Proposed Hierarchical Splatter Image}
\vspace{-2mm}

To address the two issues of the vanilla Splatter Image, we present a minimally-simple extension, as illustrated in Fig.~\ref{fig:method}, by learning more than one 3D Gaussians for each pixel, organized into a two-layer parent-child hierarchy, and by encoding the spatial information between the parent Gaussian and the target camera.  
The parent 3D Gaussians are learned in the same way as the vanilla Splatter Image. We rewrite Eqn.~\ref{eq:splatter-image} as,
\begin{equation}
    \mathcal{G}^{\text{Parent}}(I, \mathcal{P}) = g(f(I;\phi);\psi) = (\theta_1, \cdots, \theta_{H\times W}) \in \mathbb{R}^{H\times W\times M} \label{eq:parent-gaussians}
\end{equation}
where $M=24$. 

To learn child Gaussians for each parent Gaussian, we predefine the number of child Gaussians (e.g., $k=3$) to be learned. Our goal is to encode the spatial relation between the parent Gaussian and a target view. For simplicity, we encode the distance between the center of the parent Gaussian and the camera center of a target view, and their relative direction. By doing so, we expect the optimization will drive the placement of child Gaussian to be more structure aware to facilitate better syntheses of different target views. 

Let $L_{\mu}\in \mathbb{R}^{H\times W\times 3}$ be the centers/means of the parent Gaussians for each pixel, and $l_t\in \mathbb{R}^3$ the center of a target camera $t$. Both are in the world coordinate. We compute the distance map by,
\begin{equation}
    \text{Dst}(L_{\mu}, l_t) = ||L_{\mu} - l_t||_2 \in \mathbb{R}^{H\times W\times 1},
\end{equation}
and the relative direction by,
\begin{equation}
    \text{Dir}(L_{\mu}, l_t) = \frac{L_{\mu} -  l_t}{\text{Dst}(L_{\mu}, l_t) + \epsilon} \in \mathbb{R}^{H\times W\times 3},
\end{equation}
where $\epsilon$ is a small positive number to ensure numeric stability.
We introduce individual MLPs for predicting the offsets of the center of the child Gaussians with respect to the parent Gaussian, the covariance matrix, the RGB color, and the opacity. The input to the MLPs are the concatenated map $\mathcal{C}=(\mathcal{G}^{\text{Parent}}(I, \mathcal{P}), \text{Dst}(L_{\mu}, l_t), \text{Dir}(L_{\mu}, l_t))\in \mathbb{R}^{H\times W\times 28}$.  We have, 
\begin{equation}
    \mathcal{G}^{\text{child}}(I,\mathcal{P}) = (\text{MLP}_{\text{offset}}(\mathcal{C}), \text{MLP}_{\text{cov}}(\mathcal{C}), \text{MLP}_{\text{color}}(\mathcal{C}), \text{MLP}_{\text{opacity}}(\mathcal{C})) \in \mathbb{R}^{k\times (3+7+3+1)},
\end{equation}
where the MLPs consists of one hidden layer with a predefined dimension (e.g., 24) and use the ReLU as the activation function. 

We note that in order for the MLPs to have accurate sense of target view, we use target camera position in the \textit{world coordinate system} as input, instead of the relative camera position used in Splatter Image. This is important, since a strong target view sensibility helps the MLPs to accurately predict the child Gaussians to recover view-specific details. We conduct ablation studies to validate this choice in the experiment section.

\vspace{-3mm}
\section{Experiments}
\vspace{-2mm}

We evaluate our method on four datasets, including two synthetic datasets in ShapeNet-SRN~\cite{chang2015shapenet} and two real datasets in CO3D~\cite{reizenstein2021common} following the settings used in the Splatter Image~\cite{szymanowicz2023splatter}. In implementation, we build on the code of Splatter Image. \textbf{Our source code will be released too. }

\vspace{-2mm}
\subsection{Settings}\label{sec:setting}
\vspace{-2mm}

\textbf{Datasets.} We conduct evaluation on the standard single-view 3D reconstruction benchmark, ShapeNet-SRN ~\cite{sitzmann2019scene}. We train our model on single-category datasets ``Car'' and ``Chair''. We also test our method on the more challenging CO3D ~\cite{reizenstein2021common} real object datasets of ``Hydrants'' and ``Teddybear''. For CO3D, We follow the same data preprocessing procedure in the Splatter Image.

\textbf{Baselines.}
We compare our method with Splatter Image on both synthetic and real-world datasets. Additionally, for ShapeNet, we also compare against different NeRF based methods including the implicit formulations~\cite{yu2021pixelnerf,sitzmann2019scene,lin2023vision,jang2021codenerf}, hybrid implicit-explicit formulations ~\cite{gu2023nerfdiff}, and explicit formulations ~\cite{szymanowicz2023viewset,guo2022fast}; for CO3D, we also compare against PixelNeRF~\cite{yu2021pixelnerf}.

\textbf{Training Details.}
We adopt a two stage training procedure for our method: for the first 10000 iterations, we only train the U-Net architecture of Splatter Image, to ensure a reasonably good prediction of parent Gaussians; After that, we add the proposed conditional MLPs and train the entire model end to end. For training settings, we use the Adam optimizer~\cite{kingma2014adam} with batch size 8, and following Splatter Image, we add LPIPS loss after 800k iterations with weight 0.01, and train additional 100k iterations for Cars, Hydrants and Teddybears, and 200k iterations for Chairs. 
We use a single Quadro RTX 8000 GPU in experiments, and run the training for the 4 objects in our experiments on 4 GPUs respectively.

\textbf{Evaluation Procedure.}
The quality of our reconstructions is evaluated by the quality of Novel View Synthesis. We perform reconstruction from a given source view and render the 3D shape to unseen target views. We measure the reconstruction accuracy with Peak Signal-to-Noise Ratio (PSNR), and use Structural Similarity (SSIM) and perceptual quality (LPIPS) to measure structural and semantic alignment. For single-view reconstruction in ShapeNet-SRN, we follow the standard protocol: we use view 64 as the input conditioning view. For CO3D, the first frame is used as conditioning view and all remaining frames are target frames.

\begin{table*}[t]
    \centering
    
    {
    \resizebox{0.8\textwidth}{!}{
    \begin{tabular}{l|l|ccc|ccc}
    \multirow{2}{*}{Method} & \multirow{2}{*}{RC}  & \multicolumn{3}{c|}{1-view Cars} & \multicolumn{3}{c}{1-view Chairs} \\
      & &  PSNR$_\uparrow$ & SSIM$_\uparrow$ & LPIPS$_\downarrow$ & PSNR$_\uparrow$ & SSIM$_\uparrow$ & LPIPS$_\downarrow$\\ \toprule
    SRN~\cite{sitzmann2019scene} 
    & - &  22.25 & 0.88 & 0.129 & 22.89 & 0.89 & 0.104\\
    CodeNeRF~\cite{jang2021codenerf} & - & 23.80 & 0.91 & 0.128 & 23.66 & 0.90 & 0.166\\
    FE-NVS~\cite{guo2022fast}& - & 22.83 & 0.91 & 0.099& 23.21 & 0.92 & 0.077\\
    ViewsetDiff w/o $D $~\cite{szymanowicz2023viewset}& - & 23.21 & 0.90 & 0.116 & 24.16 & 0.91 & 0.088\\
    \rowcolor{blue!50} \textbf{Ours} &  -  & $+\infty^\dagger$ & \textbf{0.92} & 0.087 & \textbf{25.43}& \textbf{0.94} & \textbf{0.066}\\
    \midrule
    PixelNeRF~\cite{yu2021pixelnerf} 
    & \checkmark  & 23.17 & 0.89 & 0.146 & 23.72 & 0.90 & 0.128\\
    VisionNeRF~\cite{lin2023vision}& \checkmark &  22.88 & 0.90 & 0.084 & 24.48 & 0.92  & 0.077\\
    NeRFDiff w/o NGD~\cite{gu2023nerfdiff}& \checkmark & 23.95 & \textbf{0.92} & 0.092 & 24.80 & 0.93 & 0.070\\ 
    Splatter Image~\cite{szymanowicz2023splatter}& \checkmark & 24.00 & \textbf{0.92} & \textbf{0.078} & 24.43 & 0.93 & 0.067\\

    \midrule
    
    \rowcolor{gray!30} Splatter Image & \checkmark & 23.93 & 0.92 & \textbf{0.077}\\
    \rowcolor{blue!50} \textbf{Ours} &  -  & \textbf{24.18} & \textbf{0.92} & 0.087 \\
    
    \bottomrule
    
\end{tabular}}}
\caption{\small 
Single-view 3D reconstruction comparisons on Cars and Chairs in the ShapeNet-SRN dataset~\cite{chang2015shapenet}. 
We note that in the Car test dataset, we found that there is a pure white-background image (outlier), our method achieves a much lower reconstruction error than others, resulting in the PSNR value of $+\infty$. So, we also report results with the outlier test image excluded and compare with the vanilla Splatter Image in the last two rows.
 `{\tt RC}' means relative camera poses between input and target views are used. 
}\label{table:shapenet-single-view}
\end{table*}
\begin{figure}
    \centering
    \includegraphics[width=1.0\linewidth]{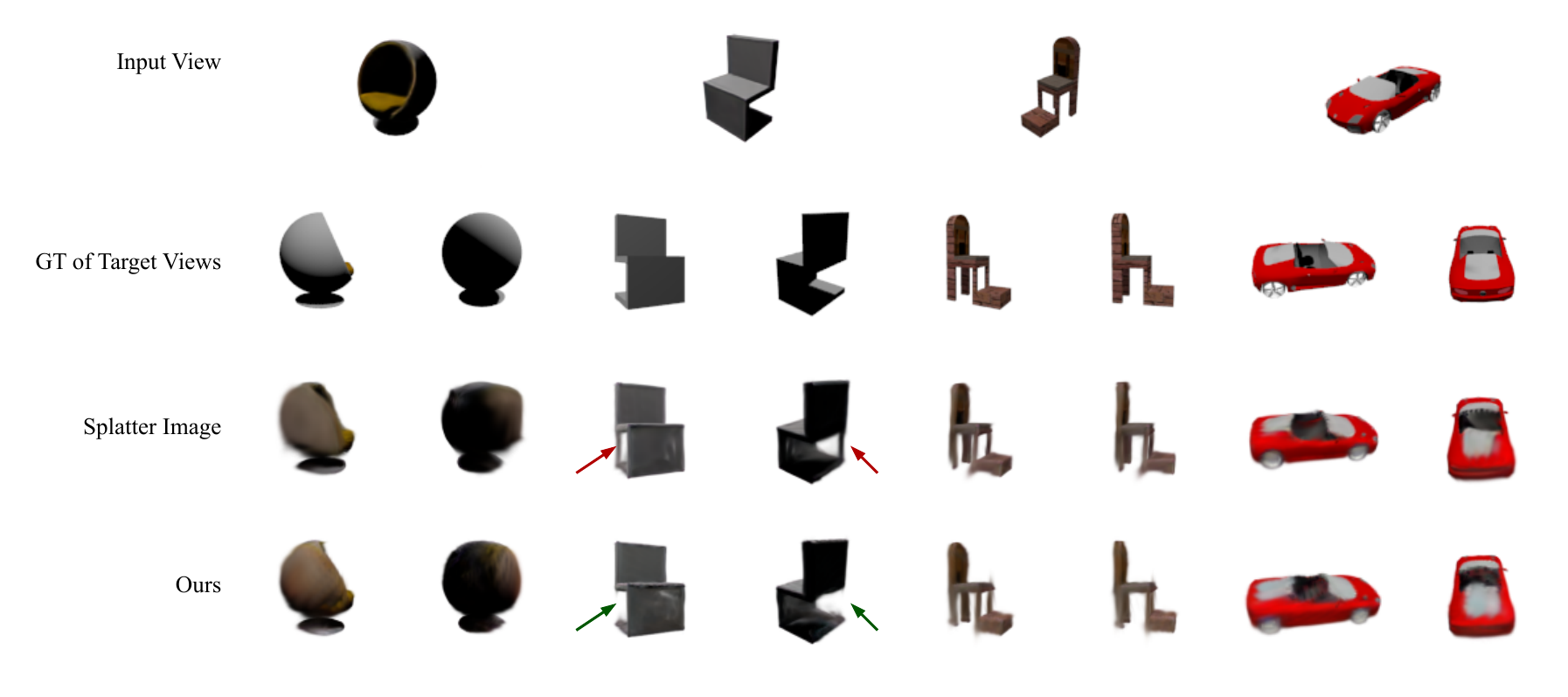}
    \caption{\small Qualitative comparisons between the vanilla Splatter Image and our proposed Hierarchical Splatter Image on Chairs and Cars in the ShapeNet-SRN test dataset.}
    \label{fig:comparison-shapenet} \vspace{-3mm}
\end{figure}

\vspace{-2mm}
\subsection{Results on Cars and Chairs in ShapeNet-SRN}
\vspace{-2mm}

\textbf{Quantitative Results.} Table~\ref{table:shapenet-single-view} shows the results. 
Overall, our method achieves the best performance across the three metrics in the Chair test dataset. It obtains the best PSNR in the Car dataset.

\textbf{Qualitative Results.} Fig.~\ref{fig:comparison-shapenet} shows examples of novel view synthesis based on single-view 3D reconstruction. Similar to the observations in Fig.~\ref{fig:teaser}, our proposed method shows more reliable novel view synthesis results in term of overall shape preservation (the first chair example and the last car example), without parts hallucinations (legs in the second chair example as indicated by the arrows) and less blurry appearance (the third chair example).

\begin{figure} [h]
    \centering
    \includegraphics[width=1.0\linewidth]{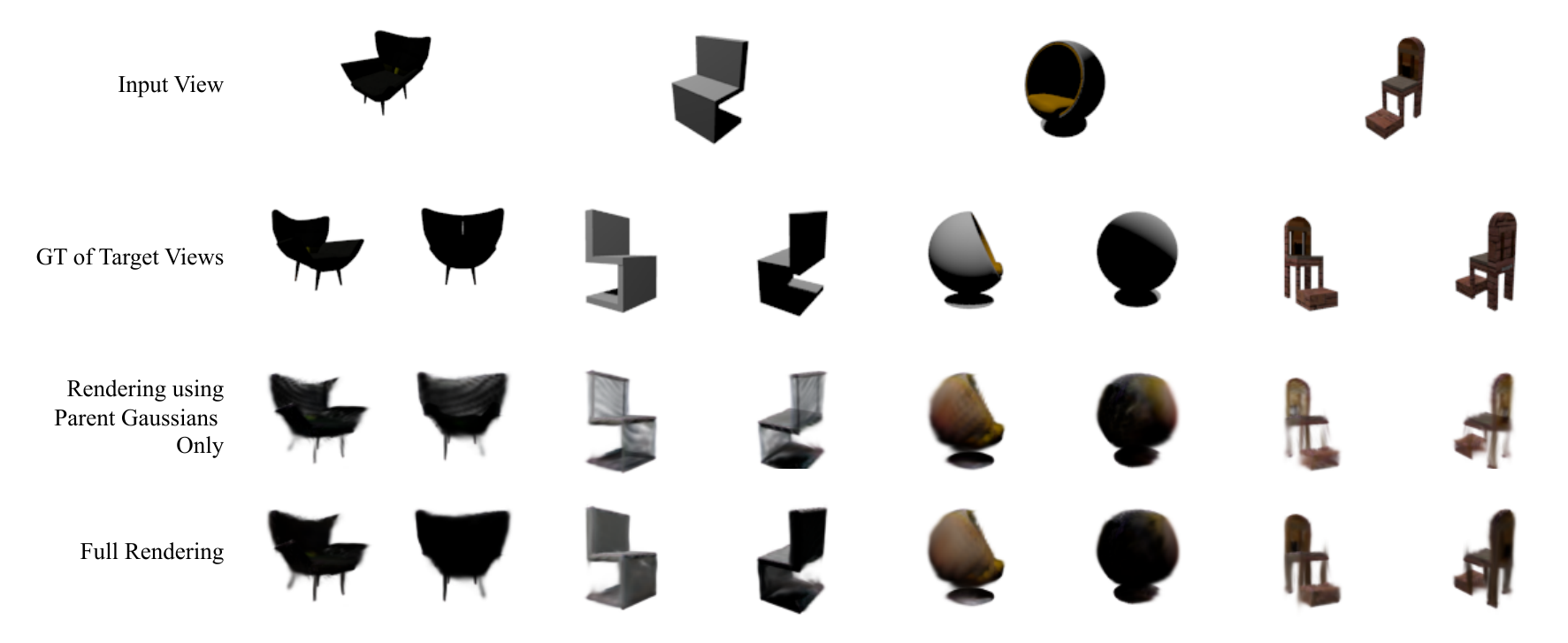}
    \caption{Visulization comparison of renders using only parent Gaussians vs full rendering with parent and child Gaussians. }
    \label{fig:comparison-parent-child-shapenet}
\end{figure}

\textbf{Parent Gaussians vs Full Gaussians.} Fig.~\ref{fig:comparison-parent-child-shapenet} shows the comparison of rendering with only the parent Gaussians vs. the full rendering. Our model learns to assign lower opacities to parent Gaussians at regions that need more detailed representation from child Gaussians (e.g., the 1st, 2nd, and 4th chair example), and keep the parent Gaussians unchanged in places that don't need refinement (the 3rd chair example). Thus, our learning mechanism mimics the effect of Gaussian splitting strategy in the original 3DGS~\cite{kerbl20233d}.

\begin{table*}
    \centering
    
    {
    \resizebox{0.48\textwidth}{!}{
    \begin{tabular}{l|c|c|c}
    \multirow{2}{*}{Method} & \multirow{2}{*}{RC}  & \multicolumn{1}{c|}{1-view Cars} & \multicolumn{1}{c}{1-view Chairs} \\
      & &  PSNR$_\uparrow$ & PSNR$_\uparrow$\\ \toprule
    
    \rowcolor{blue!50} \textbf{Ours} &  -  & \textbf{9.47}$_{+1.33}$ &  \textbf{9.89}$_{+2.39}$\\
    \midrule
    
    Splatter Image~\cite{szymanowicz2023splatter}& \checkmark & 8.14 &  7.50\\
   
    \bottomrule
    
\end{tabular}}}
~
\hfill 
{\resizebox{0.48\textwidth}{!}{
    \begin{tabular}{l|c|c|c}
    \multirow{2}{*}{Method} & \multirow{2}{*}{RC}  & \multicolumn{1}{c|}{1-view Cars} & \multicolumn{1}{c}{1-view Chairs} \\
      & &  PSNR$_\uparrow$ & PSNR$_\uparrow$\\ \toprule
    
    \textbf{Ours} &  -  & 8.40 &  8.01\\
    \midrule
    
    \rowcolor{gray!30} Splatter Image~\cite{szymanowicz2023splatter}& \checkmark & \textbf{9.13}$_{+0.63}$ &  \textbf{8.74}$_{+0.73}$\\
   
    \bottomrule
    
\end{tabular}}}
\caption{\small Close-up comparisons. \textit{Left}: pixels for which Splatter Image has high reconstruction error. \textit{Right}: pixels for which our method has high reconstruction error. By ``high'', it means the mean-square  reconstruction error (MSE) of a pixel in an image is greater than 10\% of the maximum MSE in  that image. From the margins in the subscripts, we can see that our method gains more for the ``bad pixels'' by Splatter Image than Splatter Image does for the ``bad pixels'' by our method. 
}\label{table:shapenet-single-view-forward}
\end{table*}

\textbf{Close-Up Comparisons.}  
To further compare the reconstruction performance on challenging areas within each scene, we conduct a two-way comparison. We focus on ``bad '' pixels for which Splatter Image (our method) has mean-square reconstruction error (MSE) higher than a threshold, and on which we evaluate our method (Splatter Image). Table~\ref{table:shapenet-single-view-forward} shows the results. 

Our method achieved a larger PSRN lead in Splatter Image's challenging areas, comparing to Splatter Image's lead in our method's challenging areas, demonstrating that overall our method performs better than Splatter Image in hard regions.

\begin{table*} [h]
\centering
 \resizebox{0.65\textwidth}{!}{
\begin{tabular}{l| l | c | c c c }
Object & Method & RC & PSNR$_\uparrow$ & SSIM$_\uparrow$ & LPIPS$_\downarrow$\\\hline 
\multirow{3}{*}{Hydrant} & PixelNeRF~\cite{yu2021pixelnerf}& \checkmark & 21.76 & 0.78 & 0.207 \\
 & Splatter Image~\cite{szymanowicz2023splatter}& \checkmark   &  22.10  &   0.81   & 0.148     \\
 & \cellcolor{blue!50} \textbf{Ours} & \cellcolor{blue!50} -  &  \cellcolor{blue!50}\textbf{22.45} & \cellcolor{blue!50}\textbf{0.82}    &  \cellcolor{blue!50}  \textbf{0.141}  \\
\midrule
\multirow{3}{*}{Teddybear} & PixelNeRF~\cite{yu2021pixelnerf}  & \checkmark       &   19.57    &  0.67  &  0.297 \\
 &  Splatter Image~\cite{szymanowicz2023splatter} & \checkmark   &   19.51    &  \textbf{0.73}  &   {0.236} \\
 & \cellcolor{blue!50} \textbf{Ours} &\cellcolor{blue!50} -    &   \cellcolor{blue!50}\textbf{19.70}    &  \cellcolor{blue!50}\textbf{0.73}  & \cellcolor{blue!50}\textbf{0.230}  \\
\end{tabular}}
\caption{\small Single-view 3D reconstruction comparisons on Hydrants and Teddy Bears in the CO3D dataset~\cite{reizenstein2021common}. Our method outperforms both Splatter Image and Pixel-NeRF on all metrics, except for the same SSIM on Teddybears as Splatter Image.}
\label{table:CO3D-single-view}
\end{table*}
\begin{figure} 
    \centering
    \includegraphics[width=1.0\linewidth]{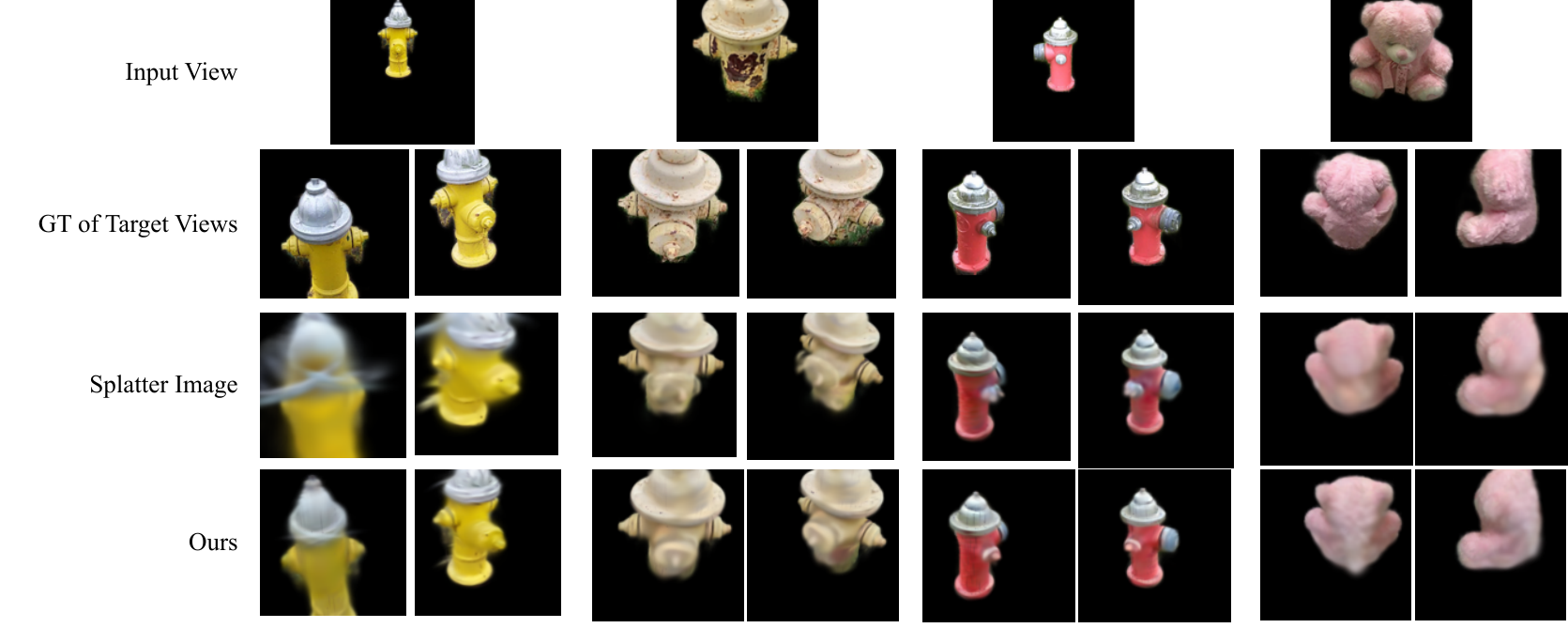}
    \caption{Qualitative comparisons between the vanilla Splatter Image and our proposed Hierarchical Splatter Image on Hydrants and Teddy Bears in the CO3D test dataset.}
    \label{fig:comparison_co3D}
\end{figure}
\subsection{Results on Hydrants and Teddy Bears in CO3D}
\vspace{-2mm}

\textbf{Quantitative Results.}
Table~\ref{table:CO3D-single-view} shows that our method consistently outperforms PixelNeRF and Splatter Image on all metrics at the challenging real world Hydrant and Teddybear datasets. On Teddybear datasets, our margin over Splatter Image is less significant, partly because instances of Teddybear don't have complex structures in the back side of the input view as those in Hydrants, as shown in Fig.~\ref{fig:comparison_co3D}.

\begin{table*}[h!] 
    \centering
    
    {
    \resizebox{0.47\textwidth}{!}{
    \begin{tabular}{l|l|c|c}
    \multirow{2}{*}{Method} & \multirow{2}{*}{RC}  & \multicolumn{1}{c|}{1-view Hydrant} & \multicolumn{1}{c}{1-view Teddybear} \\
      & &  PSNR$_\uparrow$ & PSNR$_\uparrow$\\ \toprule
    
    \rowcolor{blue!50} \textbf{Ours} &  -  & \textbf{10.87}$_{+1.36}$ &  \textbf{9.89}$_{+1.0}$\\
    \midrule
    
    Splatter Image~\cite{szymanowicz2023splatter}& \checkmark & 9.51 &  8.89\\
   
    \bottomrule
    
\end{tabular}}}
~
\hfill
{
    \resizebox{0.47\textwidth}{!}{
    \begin{tabular}{l|c|c|c}
    \multirow{2}{*}{Method} & \multirow{2}{*}{RC}  & \multicolumn{1}{c|}{1-view Hydrant} & \multicolumn{1}{c}{1-view Teddybear} \\
      & &  PSNR$_\uparrow$ & PSNR$_\uparrow$\\ \toprule
    
    \textbf{Ours} &  -  & 9.64 &  8.92\\
    \midrule
    
    \rowcolor{gray!30} Splatter Image~\cite{szymanowicz2023splatter}& \checkmark & \textbf{10.59}$_{+0.95}$ &  \textbf{9.80}$_{+0.88}$\\
   
    \bottomrule
    
\end{tabular}}}

\caption{Close-up comparisons. \textit{Left}: pixels for which Splatter Image has high reconstruction error. \textit{Right}: pixels for which our method has high reconstruction error. Pixels are selected using the same method as Table~\ref{table:shapenet-single-view-forward}.
}\label{table:co3d-single-view-forward}
\end{table*}

\textbf{Qualitative Results.}
In Fig.~\ref{fig:comparison_co3D} shows that our method can more reliably recover the unobservable structures in the input view than Splatter Image (e.g., the 1st Hydrant example), and can reconstruct  furry details on Teddybear instances better (the last column).

\textbf{Close-Up Comparisons.}
We conduct the same two-way comparison as done for Chairs and Cars in ShapNet-SRN, with the same selecting criterion. Table~\ref{table:co3d-single-view-forward} shows that our method similarly achieved larger PSNR lead at hard areas for Splatter Image, comparing to Splatter Image's lead in areas where our method have high reconstruction errors.

\vspace{-2mm}
\subsection{Ablation Studies}\label{sec:ablation}
\vspace{-2mm}

\textbf{Camera Coordinate System.} We evaluate the impact of using relative camera position and world camera position. Table~\ref{table:shapenet-single-view-ablate} shows that our method achieve consistently better PSNR than Splatter Image under both relative and world camera positions, although the margin is less significant when using relative camera poses. This is because the relative camera position of the same target view varies when the input view changes, which causes less consistent view-dependent embedding for the conditional MLPs. 
\begin{table*}[h!] 
    \centering
    
    {
    \resizebox{0.75\textwidth}{!}{
    \begin{tabular}{l|c|ccc|ccc}
    \multirow{2}{*}{Method} & \multirow{2}{*}{RC}  & \multicolumn{3}{c|}{1-view Cars} & \multicolumn{3}{c}{1-view Chairs} \\
      & &  PSNR$_\uparrow$ & SSIM$_\uparrow$ & LPIPS$_\downarrow$ & PSNR$_\uparrow$ & SSIM$_\uparrow$ & LPIPS$_\downarrow$\\ \toprule

    Splatter Image~\cite{szymanowicz2023splatter}& - & 24.12 & 0.92 & 0.089 & 25.07 & 0.93 & 0.073\\
   \rowcolor{blue!50} \textbf{Ours} &  -  & \textbf{24.18} & \textbf{0.92} & \textbf{0.087} & \textbf{25.43}& \textbf{0.94} & \textbf{0.066}\\
    \midrule

    Splatter Image~\cite{szymanowicz2023splatter}& \checkmark & 23.93 & 0.92 & 0.077 & 24.43 & 0.93 & 0.067\\ 
   \rowcolor{blue!50} \textbf{Ours} & \checkmark & \textbf{23.95} & 0.92 & 0.09 & \textbf{24.60} & 0.93 & 0.069\\
   
    \bottomrule
    
\end{tabular}}}
\caption{\small Ablation studies on camera coordinate system.} 
\label{table:shapenet-single-view-ablate} \vspace{-3mm}
\end{table*}

\textbf{Importance of encoding target views for child 3D Gaussian MLPs.}
To demonstrate that target camera view embedding is important for the conditional MLPs to predict view-specific details for better reconstruction, we conduct an ablation study, where our method is compared against a baseline that predict the child Gaussians only based on the feature of parent Gaussian, and all other configurations are kept the same. Table~\ref{table:ablate_target_camera} shows the PSNR comparison on the validation set of ShapeNet Car dataset, it can be clearly seen that without the guidance of target camera information, the baseline performs significantly worse, and doesn't improve much over the course of training. The reason of this is that without the target view as additional condition, the child Gaussian MLPs cannot accurately predict the attributes and locations of the generated Gaussians, thus resulting in even worse performance than original Splatter Image.
\begin{table}[h!]
    \centering
    \small{
    \resizebox{0.7\textwidth}{!}{
    \begin{tabular}{l|rccccccc}
    \hline 
    Model & 10k & 20k & 30k & 40k & 50k & 60k & 70k & 80k \\ \hline
    \rowcolor{blue!50} \textbf{Ours} & \textbf{20.59} & \textbf{21.42} & \textbf{21.95} & \textbf{22.32} & \textbf{22.59} & \textbf{22.78} & \textbf{22.95} & \textbf{23.10}\\
    Baseline & 19.59 & 19.61 & 19.71 & 19.79& 19.83 & 19.81 & 19.85 & 19.94\\
    \hline  
\end{tabular}  }}
\\ [1ex]
\caption{\small PSNR comparisons between our method and the baseline that does not use target camera as conditional input for child Gaussian MLPs using the ShapeNet Car validation set. }\label{table:ablate_target_camera} \vspace{-3mm}
\end{table}

\subsection{Model Complexity}
\vspace{-2mm}

\begin{wraptable}{r}{0.35\textwidth}
\centering
\resizebox{0.35\textwidth}{!}{
\begin{tabular}{l| c c  }
Method & FLOPs$_\downarrow$ & \#Params$_\downarrow$\\\hline 
Splatter Image~\cite{szymanowicz2023splatter} & \textbf{136.73G} & \textbf{56.37M} \\
\rowcolor{blue!50} Ours &  136.79G   &  61.79M    \\
\hline
\end{tabular}}
\caption{\small Model complexity comparisons.}
\label{table:complexity}
\end{wraptable}
Tab.~\ref{table:complexity} shows the model complexity comparison between our method and Splatter Image in terms of FLOPs and parameter number. Specifically, the FLOPs are measured by forward encoding process of one image. Our method has negligible increase of complexity, since the conditional child Gaussian MLPs are light weight comparing to the UNet backbone.

\vspace{-2mm}
\section{Related Work}
\vspace{-2mm}

\textbf{Neural Implicit 3D Scene Representation.} In recent years, neural implicit methods pioneered by NeRF~\cite{mildenhall2021nerf} learn to represent 3D scenes using coordinate MLPs from a set of posed multi-view images, the geometric and appearance attributes of each 3D spatial point can be extracted by querying the learned MLP with the corresponding 3D coordinates and viewing direction. Due to the expressive power of NeRF, many works extend it into generalizable inference framework, where unseen views of a scene can be synthesized from a single image. PixelNeRF~\cite{yu2021pixelnerf} leverages local latent feature extracted from input image to infer density and radiance distribution along each ray. \cite{rematas2021sharf,jang2021codenerf} parameterize MLPs with scene level latent codes, enabling generlizable geometric reasoning among similar object categories. However, the volume rendering process of NeRF requires dense queries of MLP, which is computationally expensive. 

\textbf{Voxel based 3D Scene Representation.} Another line of work ~\cite{zhao2019localization,tulsiani2017multi,peng2020convolutional,mescheder2019occupancy,kar2017learning,genova2020local,choy20163d} represent the scene explicitly with uniform feature voxels, with each voxel feature saving the geoemetric and appearance information of a spatial region, these methods significantly speed up the rendering process. However, due to the voxel grid representation, they scale poorly with resolution.

\textbf{3D Gaussian based 3D Scene Representation.} More recently, 3D Gaussian Splatting (3DGS)~\cite{kerbl20233d} achieves state of the art rendering quality in real time, it represents the scene with a set of anisotropic 3D Gaussian primitives, and learns to adjust the attributes and locations of the Gaussains to reconstruct multi-view images. The adaptive density control scheme enables the method to assign more Guassians in places that need detailed representation, while pruning Gaussains in empty space, addressing the aforementioned limitations of implicit and grid-based representations. To extend 3DGS to single / few view 3D reconstruction, Splatter Image ~\cite{szymanowicz2023splatter} proposes an image-to-image neural network that is able to predict 3D Gasussians for each pixel. However, since Splatter Image adopts a formulation that's similar to monocular depth prediction, it can only accurately predict Gaussians for pixels that are visible in the input image, and struggling to represent unseen parts of the scene with high fidelity. 

\vspace{-2mm}
\section{Limitations}\label{sec:limitation}
\vspace{-2mm}

Our proposed Hierarchical Splatter Image shows worse performance when using relative camera position, since the reference coordinate system is changing according to the input view, thus the target camera input for the MLPs is not consistent, affecting the accuracy of child Gaussian's offsets prediction. To improve our method's performance under relative camera, one possible solution is to embed richer input view information in parent Guassian's feature $\mathcal{G}^{\text{Parent}}(I, \mathcal{P})$, so that the MLPs can reason about target position better with the joint guidance of parent feature and relative camera angle. This can be achieved by aggregating multi scale features in the feature map of U-Net, through architecture like feature pyramid design~\cite{lin2017feature}.

\vspace{-2mm}
\section{Conclusion}
\vspace{-2mm}

This paper presents Hierarchical Splatter Image for single-view 3D reconstruction. It is built on the prior art, Splatter Image which in turn is built on the recently proposed 3D Gaussian Splatting (3DGS). Our proposed Hierarchical Splatter Image generalizes the one-to-one pixel-to-3D-Gaussian mapping in the vanilla Splatter Image to the one-to-many pixel-to-3D-Gaussian mapping. The one-to-many mapping is organized into a hierarchy of parent-child 3D Gaussians. A simple yet effective method is proposed for jointly learning parent and child 3D Gaussians to induce better structural awareness. The structural awareness is achieved by explicitly encoding the spatial relationships between parent 3D Gaussians and target camera poses via lightweight MLPs. Our proposed method achieves state of the art performance in four single image 3D reconstruction tasks (Chairs and Cars in the ShapeNet-SRN benchmark, and Hydrants and Teddybears in the CO3D benchmark). In particular, it reconstructs occluded contents with higher accuracy than Splatter Image.

\vspace{-2mm}
\section*{Acknowledgements}
\vspace{-2mm}
{\small J. Shen and T. Wu were partly supported by NSF IIS-1909644,  ARO Grant W911NF1810295, ARO Grant W911NF2210010, NSF IIS-1822477, NSF CMMI-2024688, NSF IUSE-2013451 and DHHS-ACL Grant 90IFDV0017-01-00.
The views and conclusions contained herein are those of the authors and should not be interpreted as necessarily representing the official policies or endorsements, either expressed or implied, of the NSF, ARO, DHHS or the U.S. Government. The U.S. Government is authorized to reproduce and distribute reprints for Governmental purposes not withstanding any copyright annotation thereon.}

\bibliography{main}
\bibliographystyle{unsrt}

\end{document}